%% file: ms.tex
\documentclass[10pt,twocolumn,letterpaper]{article}

\usepackage{cvpr}
\usepackage{times}
\usepackage{epsfig}
\usepackage{graphicx}
\usepackage{amsmath}
\usepackage{amssymb}
\usepackage{soul}
\usepackage{verbatim}
\usepackage[abs]{overpic}

\usepackage[pagebackref=true,breaklinks=true,letterpaper=true,colorlinks,bookmarks=false]{hyperref}

\cvprfinalcopy % *** Uncomment this line for the final submission

\newcommand\mypara[1]{\vspace{1mm}\noindent\textbf{#1}}

\ifcvprfinal\fi
\begin{document}

%%%%%%%%% TITLE
% \title{Realistic Evaluation of Deep Active Learning Methods}
\title{Parting with Illusions about Deep Active Learning}

\author{Sudhanshu Mittal \qquad Maxim Tatarchenko \qquad \"Ozg\"un \c{C}i\c{c}ek  \qquad Thomas Brox \\  University of Freiburg
}

\maketitle

\begin{abstract}
   Active learning aims to reduce the high labeling cost involved in training machine learning models on large datasets by efficiently labeling only the most informative samples. Recently, deep active learning has shown success on various tasks. However, the conventional evaluation scheme used for deep active learning is below par. Current methods disregard some apparent parallel work in the closely related fields. Active learning methods are quite sensitive w.r.t. changes in the training procedure like data augmentation. They improve by a large-margin when integrated with semi-supervised learning, but barely perform better than the random baseline. We re-implement various latest active learning approaches for image classification and evaluate them under more realistic settings. We further validate our findings for semantic segmentation. Based on our observations, we realistically assess the current state of the field and propose a more suitable evaluation protocol. 
   %\sudhanshu{ summarize all the suggested evaluation schemes in the conclusion section} 
   %\ozgun{I think the sentence: Current methods overlook some apparent parallel work in the literature. is too strong as it means they reinvent done methods. what we want to say instead is a parallel work in the sense of another field of research with a significant overview}
   %\ozgun{last sentence is not really correct as we do not fix a protocol.}
   
\end{abstract}

%%%%%%%%%----------------- INTRODUCTION ---------------------%%%%%%%

\input{1_introduction.tex}
\input{2_related_work.tex}

%% -----------------------------------------------

\input{3_methods_IC.tex}

\input{4_methods_SS.tex}

%% ----------------------------------------------

\input{5_discussion.tex}
%\input{latex/5_points_to_discuss.tex}

%\input{latex/6_conclusions.tex}

%% -----------------------------------------------
%\clearpage

\section*{Acknowledgements} This study was supported by the German Federal Ministry of Education and Research via the project Deep-PTL and by the Intel Network of Intelligent Systems.

\input{ms.bbl}
%{\small
%\bibliographystyle{ieee_fullname}
%\bibliography{egbib}
%}

%\clearpage
%\input{latex/7_supplementary.tex}

\end{document}

%% file: 1_introduction.tex
\section{Introduction}

%Supervised learning models have shown great deal of success in solving various visual tasks.
Supervised training of convolutional networks has shown remarkable success in various computer vision tasks.
Its price is the collection of large datasets and their annotation.
Especially the data annotation is a common bottleneck. 
Depending on the task, its cost may vary from a few seconds to a few hours per sample.
%involving a layperson to an domain expert.
%Thus, creating large datasets can easily become prohibitively expensive for many practical applications.

Active learning (AL) presumably mitigates this large annotation cost.
% The goal of AL is to achieve maximum performance with minimum annotation cost.
It is based on the attractive idea that some samples are more valuable for learning than others - by identifying those in the pool of unlabeled data, we can use an annotator's time more efficiently.
%Conventionally, AL methods rely on the assumption the most useful bias comes from training on the most difficult samples.
The typical process in AL includes multiple cycles, where in each cycle a batch of samples is selected from the pool of still unlabeled data using a query function.
The selected samples are manually annotated and are added to the labeled set. Then the model is re-trained.
The process is repeated until the maximum annotation budget or the desired performance level is reached.
% In practice, AL methods select and label the most informative samples such that the best performance is achieved given a fixed annotation budget or a certain performance is achieved with minimum annotation cost. Active learning is an iterative process where in each cycle a batch of samples are selected from a unlabeled pool based on a query function. Active learning, especially for deep neural network, involves labeling new batch of samples since adding single sample is insignificant for large deep networks.

%\ozgun{for figure caption we do not really have a recommended scheme do we?}

\begin{figure}[t]
    \includegraphics[width=\columnwidth]{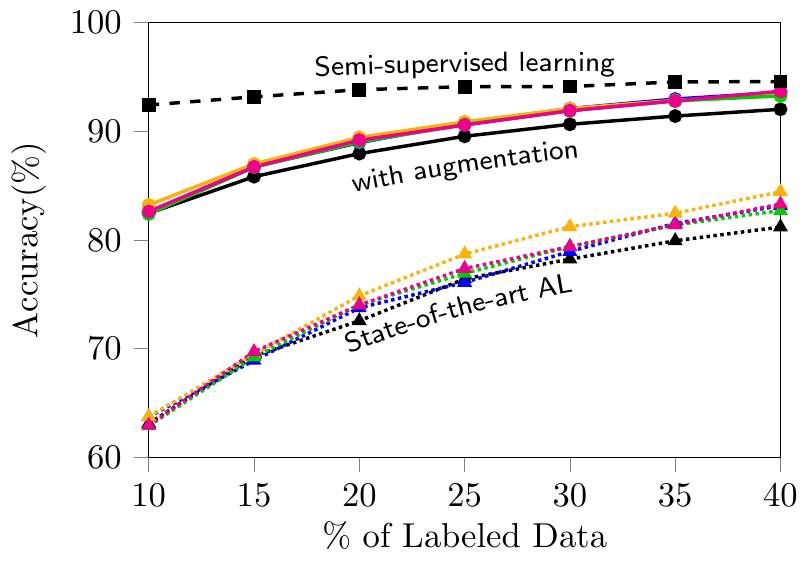}
        \caption{State-of-the-art active learning methods do not consistently use modern data augmentation techniques or advances in the closely related field of semi-supervised learning which leads to the wrong impression about the current state of the field.
        %Improvement over the state of the art in active learning when proper augmentation is used, and when the same unlabeled data pool is rather used for semi-supervised learning (dashed black curve, no active learning).
        Results are shown for image classification on CIFAR 10.}
    \label{fig:c10_teaser}
\end{figure}

%\maxim{\st{This paragraph can be replaced with two general sentences about what AL does here in the intro. A detailed description can come later.}}

%\maxim{\st{I would be more aggressive here, otherwise we may run into the same problem as the concurrent ICLR paper. We can explicitly say that SSL and AL use the same amount of information but are currently often treated as unrelated methods. Also, AL models are deliberately weakened by not using appropriate data augmentation.}}

The appeal of the active learning idea has spawned a multitude of ConvNet-based AL methods.
In this paper we aim to objectively assess the state of the field and  challenge the principal hypothesis behind active learning: active selection of the samples to be labeled leads to a significant reduction in the annotation effort compared to random selection. 
% The challenge of active learning has lately drawn a lot of attention since it can potentially address the rising need of large labeled dataset for various practical applications.
Our study seeks answers to the following four scientific questions.

%We find that 
%the current evaluation scheme used in AL works is unrealistic and fails to adequately characterize the methods' performance.
(1) Since a widely accepted evaluation protocol is missing, methods are often tested under incompatible circumstances: different architectures, different augmentation strategies, etc.
We evaluate the effect of compatible experimental settings on the ranking of methods.
In particular, do  AL  methods  work  consistently  well  in conjunction with data augmentation?
% Current works lack a standardized evaluation protocol and tend to test methods in different configurations. 
%Methods are tested with different architectures, augmentation techniques and annotation budget, thus leading to discrepancies in reported results. 
%Semi-supervised learning share common goals with active learning and utilize same amount of resources to attain much higher performance.

(2) Contemporary papers on active learning largely ignore the progress of the closely related field of semi-supervised learning, where approaches effectively operate under the same assumptions with regard to the used data. What is the effect as concepts from semi-supervised learning are integrated into active learning?
% Current AL works ignore the progress of concurrent research work in semi-supervised learning and are often treated as unrelated methods, although they share common goals.
% Also, we find AL models are deliberately weakened by not using appropriate data augmentation.

(3) Existing methods are typically not evaluated in a low-budget setting - a mode crucially important to kick-start network training on a new dataset. How do active learning concepts work in such low-budget regime? 
% A further study into a low-budget setting is usually missing, which could be crucial for expensive labeling tasks and to start the labeling process early. Based on these aforementioned issues, we raise following critical questions:

(4) Active learning is typically evaluated only on image classification tasks where manual labeling is relatively cheap compared to other tasks, \eg semantic segmentation. Can active learning better exploit its potential in the more costly scope of semantic segmentation, where efficient annotation is practically more relevant?

%\sudhanshu{Teaser paragraph??}
In this work, keeping in mind the aforementioned questions, we perform an extensive comparison of existing approaches for image classification and semantic segmentation. Our experiments reveal that the progress recently made in the field of active learning is practically negligible when viewed under more realistic circumstances: in particular, using modern data augmentation and taking the advances of semi-supervised learning into account, see Figure \ref{fig:c10_teaser}.
Based on our extensive study, we suggest a more suitable evaluation protocol. 

%% file: 2_related_work.tex
\section{Related Work}

%\ozgun{overall the related work [x] is used as subject in the sentences which should be fixed for better readability.}

\mypara{Deep active learning (AL)} methods can be categorized into three types: uncertainty-based methods, representation-based methods and learning-based methods. Additionally, some methods have proposed a hybrid approach. 

%VERSION 1:
%Uncertainty-based methods try to find the samples which are hard to learn. Several methods have been proposed to estimate uncertainty for neural networks using dropouts \cite{Gal2016Bayesian, Gal:2016:DBA:3045390.3045502, Gal:2017:DBA:3305381.3305504},  highest entropy \cite{5206627, Wang:2017:CAL:3203306.3203314}, distance from the decision boundary \cite{Tong:2002:SVM:944790.944793} and degree of committee disagreement \cite{include query by committee}. 

% Version 2: 
\textit{Uncertainty-based methods} try to find the samples which are hard to learn. Several methods have been proposed to estimate uncertainty for neural networks using bayesian \cite{Blundell:2015:WUN:3045118.3045290, Gal2016Bayesian, Gal:2016:DBA:3045390.3045502, visionUncertainties} and non-bayesian approaches \cite{NIPS2017_7219, NIPS2016_6501}. Gal \etal \cite{Gal:2017:DBA:3305381.3305504} proposed to estimate posterior uncertainty using dropout for active learning. Wang \etal \cite{Wang:2017:CAL:3203306.3203314} used the entropy of the softmax output in a neural network as a proxy uncertainty measure to query samples. Beluch \etal \cite{Beluch_2018_CVPR} use ensemble method to estimate prediction uncertainty and selects new samples based on a statistical measure of committee disagreement called variation ratio \cite{10.1093/sf/44.3.455}. They show this method outperforms all other uncertainty-based methods. 
\textit{Representation-based methods} \cite{ sener2018active,SA}, also referred as density-based methods, try to find a diverse set of samples that optimally represents the complete dataset distribution. Sener \etal \cite{sener2018active} formulated the active learning problem as core-set selection and showed effectiveness for CNNs.  
\textit{Learning-based approaches} \cite{Sinha_2019_ICCV, Yoo_2019_CVPR} use an auxiliary network module and loss function to learn a measure of information gain from new samples. Yoo \etal \cite{Yoo_2019_CVPR} proposed to learn a loss prediction module to predict target losses of unlabeled samples and selects samples with highest predicted loss. It can also be considered as a pseudo-uncertainty heuristic. 
Sinha \etal \cite{Sinha_2019_ICCV} proposed a semi-supervised active learning approach that learns a VAE-GAN hybrid network to select unlabeled samples that are not well represented in the labeled set. It can also be considered as a representation type method. 

% \textit{AL for Semantic Segmentation}
Many of the above mentioned approaches mainly focus on image classification. Lately, a few works have proposed to solve tasks involving higher annotation cost like object detection \cite{Yoo_2019_CVPR}, pedestrian detection \cite{Yoo_2019_CVPR}, human pose estimation \cite{8237730} and segmentation \cite{7780682, Sinha_2019_ICCV}. We focus on semantic segmentation in this work, since creating segmentation masks is a highly expensive labeling task. This makes it one of the most relevant task for active learning. Suggestive Annotation \cite{SA}, Cereals \cite{Mackowiak2018CEREALSC} and VAAL \cite{Sinha_2019_ICCV} are few works which have shown applicability of deep active learning for semantic segmentation. VAAL is a task-agnostic learning-based approach using adversarial training. Suggestive Annotation is a hybrid approach proposed for a binary segmentation problem. Cereals is a patch-based selection approach based on a hybrid heuristic of uncertainty, learned labeling cost model and spatial coherency of the image. 
 
\mypara{Semi-supervised Learning (SSL)} methods make use of the unlabeled data for training the model. Effectively, this class of methods uses the same amount of information as the AL methods. SSL has recently seen a lot of progress due to consistency regularization \cite{mixmatch,  Xie2019UnsupervisedDA, Zhai_2019_ICCV}. Consistency regularization minimizes the discrepancy between class predictions of differently perturbed unlabeled image. Various additional schemes have been proposed to avoid overfitting and to improve training stability, such as temporal ensembling \cite{conf/iclr/LaineA17}, student-teacher model \cite{ NIPS2017_6719}, adversarial perturbation \cite{8417973}, self-supervision \cite{Zhai_2019_ICCV}, data filtering \cite{Xie2019UnsupervisedDA}, and snapshot ensembling \cite{athiwaratkun2018there}. In this work, we use a semi-supervised learning method UDA \cite{Xie2019UnsupervisedDA} for image classification.

A few recent works \cite{Hung_semiseg_2018, 1908.05724} applied semi-supervised methods also to semantic segmentation. Mittal \etal~\cite{1908.05724} proposed s4GAN that uses a conditional GAN \cite{NIPS2014_5423} to learn from unlabeled images,  French \etal \cite{DBLP:journals/corr/abs-1906-01916} used consistency regularization and Kalluri \etal \cite{Kalluri_2019_ICCV} proposed a feature alignment objective to learning from unlabeled samples. We make use of s4GAN \cite{1908.05724} in this work.

\mypara{Semi-supervised Active Learning}. Most representation-based AL methods use unlabeled samples to learn the underlying distribution, but only a few methods use semi-supervised learning to improve their selection criteria \cite{ravanbakhsh:MIDLAbstract2019a, sener2018active, Sinha_2019_ICCV, Wang:2017:CAL:3203306.3203314}. Sinha~\etal~\cite{Sinha_2019_ICCV} used unlabeled pool to learn its distribution against the distribution of labeled samples, but do not take its advantage to improve the feature representation of the target model itself. Sener~\etal~\cite{sener2018active} have also previously shown the advantage of using the unlabeled pool for learning the target model. Wang~\etal~\cite{Wang:2017:CAL:3203306.3203314} also explored the usage of the most-certain samples from the unlabeled pool using pseudo-labeling, but the pseudo-labeling process can easily propagate erroneous labels if not tuned properly. Ravanbakhsh~\etal~\cite{ravanbakhsh:MIDLAbstract2019a} proposed a GAN-based approach to make use of the unlabelled pool and utilizes the discriminator score to query low-confident samples for active learning.  Recently, two open-source concurrent works \cite{ssl_al, rethinking_al} have also shown some similar findings as our work. However, they are restricted to only image classification.  

Oliver~\etal~\cite{NIPS2018_7585} raised concerns about the evaluation scheme for semi-supervised learning to help guide its applicability to real-world problems. In this work, we raise similar questions about unrealistic evaluation schemes for active learning by providing evidence through extensive experiments. 
In the following sections, we analyze the performance of active learning methods for image classification and semantic segmentation respectively. 

%labeling cost for these two applications differs significantly. While labeling each image for classification requires 1 click per image, labeling an image for semantic segmentation costs manifold.

%% file: 3_methods_IC.tex
\section{Active Learning for Image Classification} \label{Sec:IC}

In active learning, we usually start with a small set of labeled samples $\mathcal{L}$ whose size is defined by the initial labeling budget $\mathcal{B}_i$ and a large pool of unlabeled samples $ \mathcal{U}$.
In each cycle, a set of samples is selected from the unlabeled pool $\mathcal{U}$ according to the sampling budget $\mathcal{B}_s$ and added to the labeled set $\mathcal{L}$ with the corresponding annotations provided by the oracle annotator. 
The selection of samples is performed using a query function, which can be learned using the full set of available samples ($\mathcal{U} \cup \mathcal{L}$). This process is also called pool-based active learning. This acquisition step is iterated over several cycles until the objective is achieved.

% ------------- METHODS -------------------------------------
In this section, we assess the performance of state-of-the-art AL methods for image classification and compare them with the state-of-the-art semi-supervised approach. We validate our experiments using at least one recent approach from each of three categories of AL methods as defined in the related work section.

\subsection{Baseline Methods}

\mypara{Random.} A new set of samples is selected randomly from the unlabeled pool and is added to the labeled pool with annotations.

\mypara{Entropy} \cite{Shannon1948} is an information-theoretic measure used as an uncertainty metric for sampling.
This method naively selects samples for which the pseudo-probabilities predicted by the softmax classifier have the highest entropy.

%calculated as follows: 
%\begin{align}
%    H(y|x_u) = -\sum_c(p(y=c|x_u))\log(p(y=c|x_u)).
%\end{align}
% For the entropy method, we use the softmax output of the final fully-connected layer to calculate the entropy of the prediction.

\mypara{Ensemble with Variation Ratio (ENS-varR).} The second method, which selects samples based on an uncertainty criterion relies on using ensembles. It has been shown to consistently outperform all other uncertainty-based approaches for active learning by Beluch~\etal~\cite{Beluch_2018_CVPR}. 
%\ozgun{in uncertainty or al? I find this method explanation a bit confusing}
% Specifically, the Variation Ratio (varR) metric in combination with ensemble model shows improved performance.
%\maxim{What is a modal class?}
The core of the method is to calculate the variation ratio (varR) metric given as the proportion of predicted class labels that are not the modal class prediction:
\begin{align}
    varR = 1 - \frac{f_m}{T},
\end{align}
where $f_m$ is the frequency of the modal class and $T$ is the number of ensemble members. This heuristic is motivated by the query-by-committee algorithm proposed by Seung~\etal~\cite{Seung:1992:QC:130385.130417}.
The query function selects the samples with larger varR values.
The ensemble is only used for sample querying - the target performance is still reported for a single model.
Similar to Beluch \etal \cite{Beluch_2018_CVPR}, we use an ensemble of 5 models for our experiments.
% For ensemble-based method, to keep the memory resources tractable and comparable to other methods, we exploit the ensemble model to calculate the uncertainty of the sample prediction and only use a single model to evaluate the performance of the target model. 

\mypara{Core-set}. This type of method selects a batch of samples such that the performance of the model trained on the labeled set matches the performance of the model trained on the whole dataset~\cite{6907021}. The recent core-set approach proposed by Sener~\etal~\cite{sener2018active} casts the core-set selection problem as a k-center problem and proposes a robust k-center approach. The proposed approach chooses a subset, such that the largest distance between chosen point and unlabeled points is minimized in the feature space. For the core-set approach, we make use of the k-center greedy implementation since it is much faster and only performs marginally worse than the robust version.

\mypara{Learning Loss (LL)}. This method \cite{Yoo_2019_CVPR} proposes a loss prediction module which is attached to the target network to estimate the loss value of the unlabeled samples. The samples with the largest predicted loss are selected for annotation.
This auxiliary module is trained to preserve the pairwise ranking of the original loss values which is imposed using a hinge loss function over random pairs of samples in a minibatch.
% This auxiliary module is trained  by regularizing the order of predicted loss values to remain same as the original loss values. This is imposed using a hinge loss function over random pair of samples in a minibatch.

%The unlabeled samples with the highest predicted loss value are considered to be the most informative for the learning process and are thus selected by the query function for labeling.

\mypara{Unsupervised Data Augmentation (UDA).}
%\maxim{\st{A bit more introduction is required into why we take this particular method.}}
UDA \cite{Xie2019UnsupervisedDA} is a semi-supervised learning method for image classification. It uses consistency regularization to learn from unlabeled samples along with AutoAugment \cite{Cubuk_2019_CVPR} and other augmentation techniques to reduce overfitting. We selected this method because: 1) it shows state-of-the-art performance, 2) it is based on a simple idea and is easy to implement. Also, the method performs well even when the number of labeled samples is very small. Our implementation used online data augmentation instead of the offline one in the original work \cite{Xie2019UnsupervisedDA}.

%-------------- EXPERIMENTS -------------------------

\subsection{Experiments and Results}

\subsubsection{Evaluation protocol}

%\maxim{These setups are confusing here. How about only specifying the large-budget setting at this point, and then just separately talking about the low-budget setting in the corresponding section?}

\textbf{Datasets}. We evaluate the methods on the CIFAR-10 and CIFAR-100 datasets. Both datasets contain the same set of 60,000 images, assigned to 10 and 100 classes respectively. The training and test set contain 50,000 and 10,000 images respectively. CIFAR-10 is the most commonly tested dataset in the field of active learning. CIFAR-100 is an extension with 100 classes, which makes the task more challenging. The initial labeling budget is $\mathcal{B}_i=5000$ and the sampling budget is $\mathcal{B}_s=2500$ labels for each cycle.
We tested this configuration for 6 sampling cycles (\textit{i.e.} going from 10\% to  40\% labeled samples). In the first step, we randomly sampled a class-balanced subset of samples from the unlabeled pool.
 
\textbf{Training Details}. For the network architecture, we consistently use the Wide-Resnet-Network \cite{Zagoruyko2016WRN} with depth=28 and width=2 (WRN-28-2). We select WRN due to its efficiency and widespread adoption. WRN-28-2 contains only 1.5M parameters showing close-to-state-of-the-art performance on CIFAR datasets. 
The WRN-28-2 classification network is optimized using SGD optimizer with a base learning rate of 3e-2, momentum 0.9 and weight decay rate of 5e-4. We use a cosine learning rate schedule for training each model.
We trained all AL methods (without SSL methods) for 150 epochs per sampling cycle with a batch size of 64.  
We train the semi-supervised AL methods for 50k iterations 
% (recommended by Xie \etal~\cite{Xie2019UnsupervisedDA}: 400k for best performance)
per sampling cycle with a batch size of 64 for the labeled loss and a batch size of 320 for the unlabeled loss. We mask out unlabeled examples whose highest probabilities across categories are less than 0.6 and set the softmax-temperature scaling constant to 0.5. Other hyperparameters are used exactly as proposed in \cite{Xie2019UnsupervisedDA}.
Our implementation is based on the open source toolbox Pytorch \cite{paszke2017automatic}.

\begin{figure}[t]
    \includegraphics[width=\columnwidth]{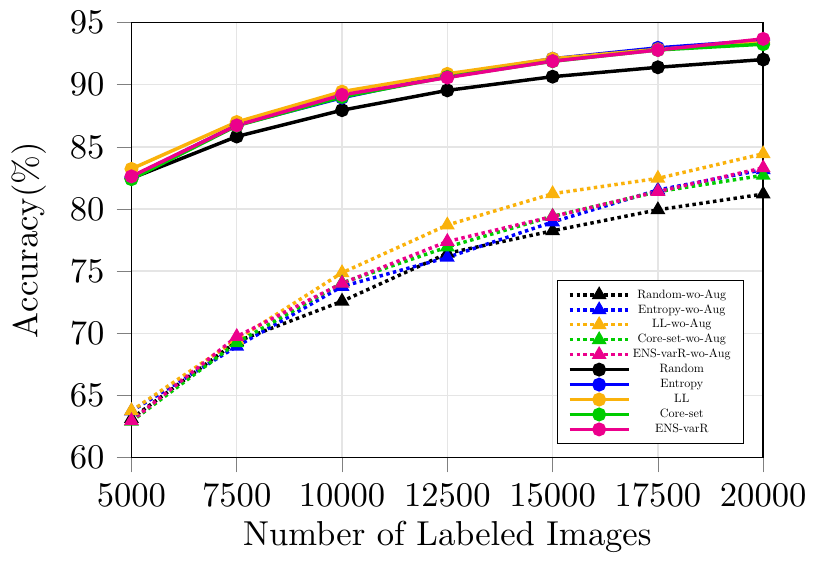}
       \caption{Using data augmentation on CIFAR-10 significantly improves the performance of active learning methods and makes the relative difference between them less pronounced. Results without augmentation are denoted as 'X-wo-Aug'.
       }
    \label{fig:w_vs_wo_aug}
    %\vspace{-2mm}
\end{figure}

All results are shown as performance curves. We report the mean performance over 3 trials with different initial labeled sets for all single model-based methods and over 2 trials for ensemble-based methods due to higher computation cost and lower variance.

LL methods usually starts with a higher initial performance due to the extra regularization effect from the loss-prediction module. All other methods start from similar initial performance with slight difference due to the model variance. This variance is more prominent in the beginning due to the overfitting effect on small labeled set.

% ---------------- Section 3.2.2 --------------------------

\subsubsection{Do AL methods work consistently well together with data augmentation?}
\label{Sec:IC_w_aug}

Data augmentation is a widely accepted regularization technique, which increases the power of machine learning models particularly when there is little labeled data.
Nevertheless, several latest AL works \cite{Beluch_2018_CVPR, Sinha_2019_ICCV} resort to either not using any augmentation during training, or only doing simplistic horizontal flipping.
In this experiment, we validated the importance of elaborate up-to-date image augmentation for the performance of AL methods.
% Here, we propose to make use of the more involved augmentation techniques and observe how well different AL methods perform with augmentations. 

We first evaluated all methods without any augmentation. Subsequently, we evaluated the same methods with augmentation, which includes using the AutoAugment policies found by Cubuk~\etal~\cite{Cubuk_2019_CVPR} , cutout \cite{devries2017cutout}, horizontal random flipping, and random cropping.
Figure \ref{fig:w_vs_wo_aug} shows that without using any augmentation, all AL methods clearly perform better than the random baseline. 
%Results for CIFAR-100 can be found in the supplemental material.
The LL method shows distinct improvement over other methods (matching the results from Yoo~\etal~\cite{Yoo_2019_CVPR}) and an overall improvement of 3.2\% over the random baseline on the CIFAR-10 dataset.
When the same experiment is performed with augmentation, all the methods improve drastically in absolute performance.
However, the relative effect of using different AL methods becomes far less pronounced: all the AL methods show similar performance within a range of 0.4\%.
In conclusion, AL works well with data augmentation, but data augmentation blurs the differences between AL strategies: they all perform largely the same. 

\begin{figure}[t]
    \includegraphics[width=\columnwidth]{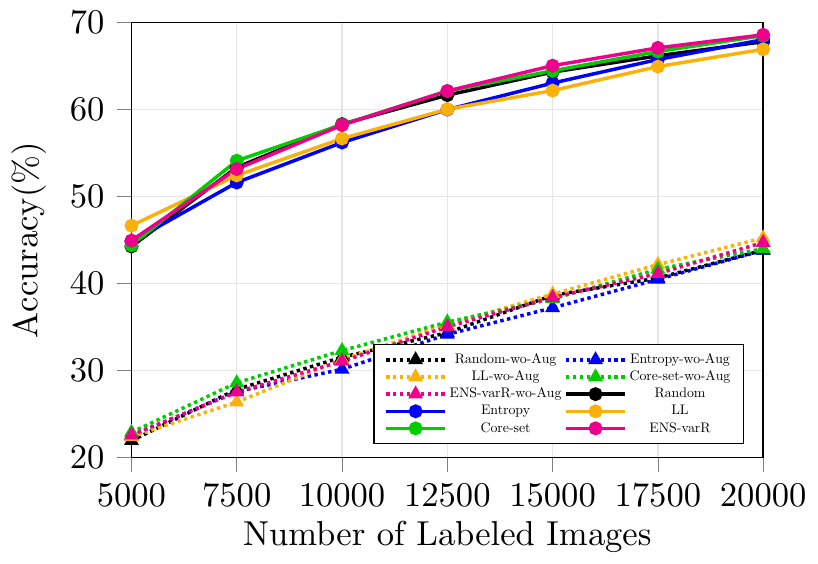}
    \caption{The performance of AL methods on CIFAR-100 improves significantly when using up-to-date image augmentation. Results without augmentation are denoted as 'X-wo-Aug'.}
    \label{fig:c100_w_vs_wo_aug}
    %\vspace{-8mm}
\end{figure}

For completion, we further validate the importance of using up-to-date augmentation for AL methods on the CIFAR-100 dataset. We evaluate all methods with and without augmentation similar to the CIFAR-10 experiment. The overall conclusion is also very similar: Without augmentation, the LL method shows distinct improvement of 1.4\% over the random baseline; with augmentation, all the methods improve by a large margin in absolute performance but the relative difference between different methods becomes insignificant and the relative ranking of different methods changes. Performance curves are shown in Figure \ref{fig:c100_w_vs_wo_aug}. % and the exact performance numbers are shown in Table \ref{table:c100_w_vs_wo_aug}.

\subsubsection{Does semi-supervised learning or active learning make better use of the pool of unlabeled data?} \label{Sec:IC_SSL}

A largely common practice in the previous works has been to utilize the unlabeled pool only for sampling, although it is available throughout the learning process (otherwise one could not sample from it) and could be used more rigorously.
Using semi-supervised learning, we can utilize this unlabeled pool for training the model itself. To this end, we employed the UDA semi-supervised learning method.
We integrated SSL into the AL methods by training the model using the UDA objective and defining the query function based on this model. 
In each cycle, the target model is trained using UDA instead of the standard supervised training.
Data augmentation stays the same as in Sec.~\ref{Sec:IC_w_aug}.
We refer to the integrated methods as SSL-X, where X is the name of the AL method.

\begin{figure}[t]
    \includegraphics[width=\columnwidth]{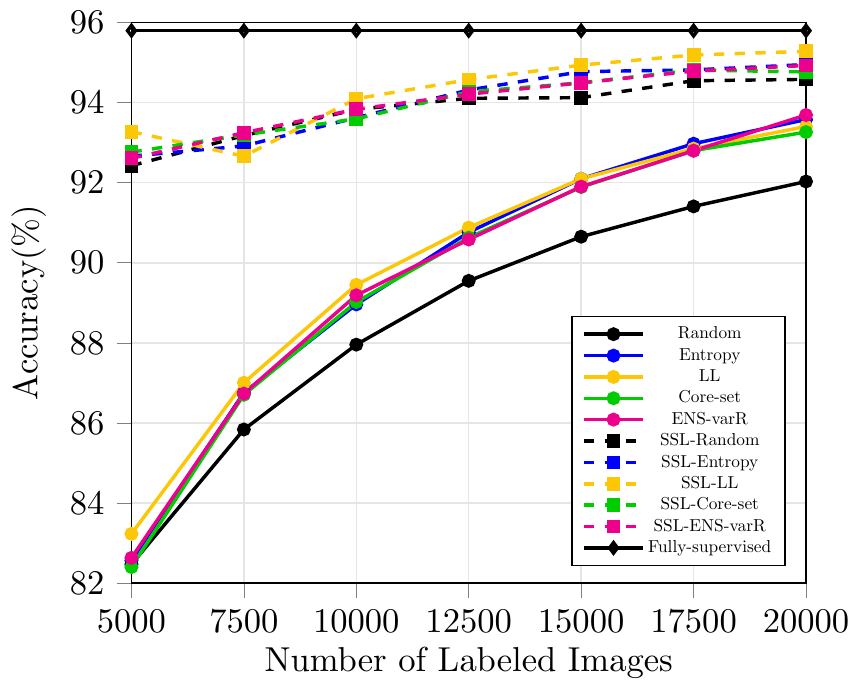}
       \caption{Combining AL methods with semi-supervised learning leads to significant performance improvement on CIFAR-10 compared to the raw AL case. Results shown in the large-budget setting with $\mathcal{B}_i=5000, \mathcal{B}_s=2500$.}
       % Results on CIFAR-10 comparing SSL integrated active learning (SSL-X) methods with raw active learning methods. Results in low-budget setting, where $\mathcal{B}_i=5000, \mathcal{B}_s=2500$.} 
    \label{fig:c10_5k}
\end{figure}

\begin{figure}[t]
    \includegraphics[width=\columnwidth]{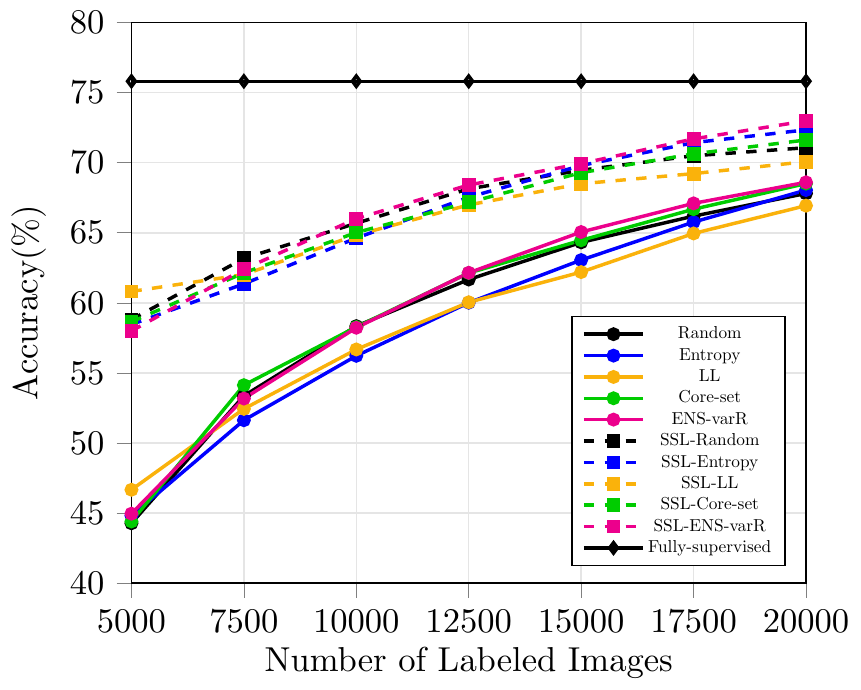}
       \caption{Integrating SSL and AL leads to overall performance improvement on CIFAR-100, however, not all combinations consistently outperform random sampling. Results shown in the large-budget setting with $\mathcal{B}_i=5000, \mathcal{B}_s=2500$.}
       % Results on CIFAR-100 comparing SSL integrated active learning (SSL-X) methods with raw active learning methods. Results in low-budget setting, where $\mathcal{B}_i=5000, \mathcal{B}_s=2500$.}
    \label{fig:c100_5k}
\end{figure}

Figures~\ref{fig:c10_5k} and \ref{fig:c100_5k} show a remarkably strong performance of the SSL method (SSL-Random) on CIFAR10 and CIFAR100: when using 5K random labeled samples, SSL almost reaches the same performance which AL methods achieved on 20K samples picked by the corresponding query functions. 
Also for the remaining data ratios, there is a large performance gap between semi-supervised and active learning, both on CIFAR-10 and CIFAR-100. 
Clearly, semi-supervised learning makes much better use of the same data than active learning. 
% When the unlabeled pool is used to train the model, we observe a substantial improvement over previous settings used in AL works, even over the one with more involved augmentations, see Figure \ref{fig:c10_5k}.

SSL and AL can be combined, which yields an improvement over raw SSL on CIFAR-10. The SSL-LL method performs best and shows an improvement over the random baseline by 0.7\% after 6 cycles. However, on CIFAR-100 the relative ranking of the AL methods changes completely; SSL-LL performs worse than the other methods and struggles even to compete with the random selection method. 

The same is true for raw active learning without SSL: on CIFAR-100 some active learning methods do not reach the performance of randomly drawing the samples to be labeled, shown in Figure \ref{fig:c100_5k}. 
%Here, SSL-ENS-varR outperforms all other methods, while the simple SSL-Entropy method is second-best, leaving behind more sophisticated AL methods.

%Although the performance of integrated methods are slightly inconsistent in order, the importance of using integrated methods is clearly justified by the huge performance boost over the 'supervised only' baseline methods. Therefore, this integration of SSL should be made a standard protocol when proposing a query method for active learning.

%-------------------- Section 3.2.4  -------------------------

% \subsubsection{Low-budget Setting} \label{Sec:IC_w_low}
\subsubsection{Does active learning consistently outperform random sampling in low-budget regimes?} \label{Sec:IC_w_low}

%The above finding is that a simple change of the dataset from CIFAR-10 to CIFAR-100 challenges the principal assumption of active learning, that a dedicated selection strategy improves over random selection of samples.
There is an inconsistency in the methods' behavior when switching from CIFAR-10 to CIFAR-100. This challenges the principal assumption of active learning that a dedicated selection strategy always improves over random selection of samples.
Does active learning benefit from a low-budget setting, where every sample is particularly crucial?
% For certain applications like bio-medical image analysis, class labels can be expensive therefore the total labeling budget can be small.
In certain applications, such as medical image analysis, already 10000 annotated samples can be very costly. 
Thus, training with only few labeled samples in the beginning is attractive.
%Further, it can be crucial to start the active learning process early. This challenge is also referred as the cold-start problem, also addressed in \cite{pmlr-v32-houlsby14} for recommendation systems.
We explored such low-budget setting with $\mathcal{B}_i$ and $\mathcal{B}_s$ for each cycle set to 250 labels for CIFAR-10 and 500 labels for CIFAR-100. We tested this setting for 7 sampling cycles with a total budget of 2000 and 4000 labels for CIFAR-10 and CIFAR-100, respectively.
We kept all the augmentation techniques from the previous experiments.

\begin{figure}[t]
    \includegraphics[width=\columnwidth]{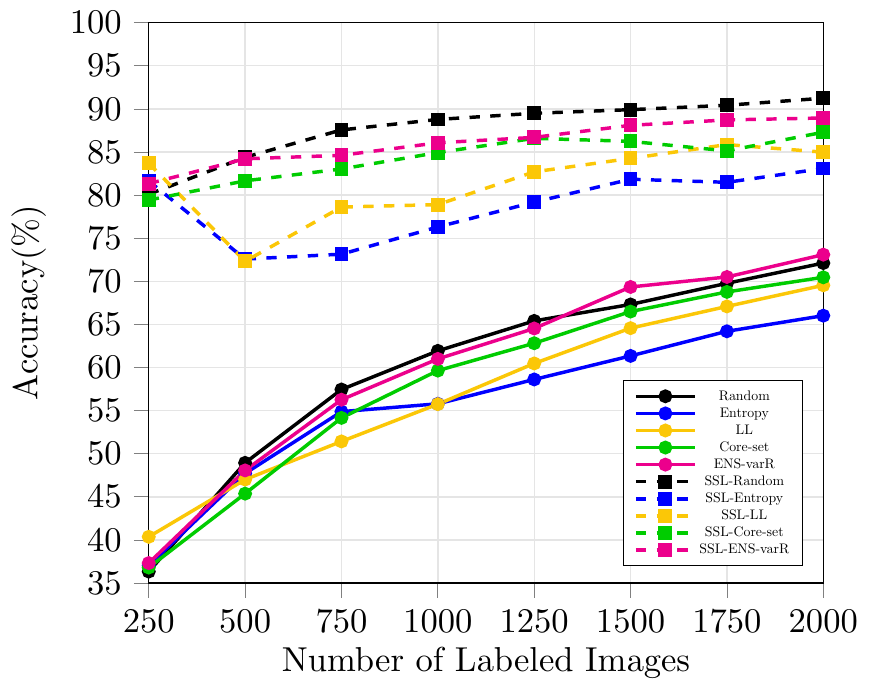}
       \caption{When evaluated in the low-budget regime ($\mathcal{B}_i=\mathcal{B}_s=250$) on CIFAR-10, integrated SSL-AL methods are still better than their raw counterparts, however, SSL with random sampling shows the best performance.}
       % Results on CIFAR-10 comparing SSL integrated active learning (SSL-X) methods with raw active learning methods. Results in low-budget setting, where $\mathcal{B}_i=\mathcal{B}_s=250$.}
    \label{fig:c10_250}
\end{figure}

\begin{figure}[t]
    \includegraphics[width=\columnwidth]{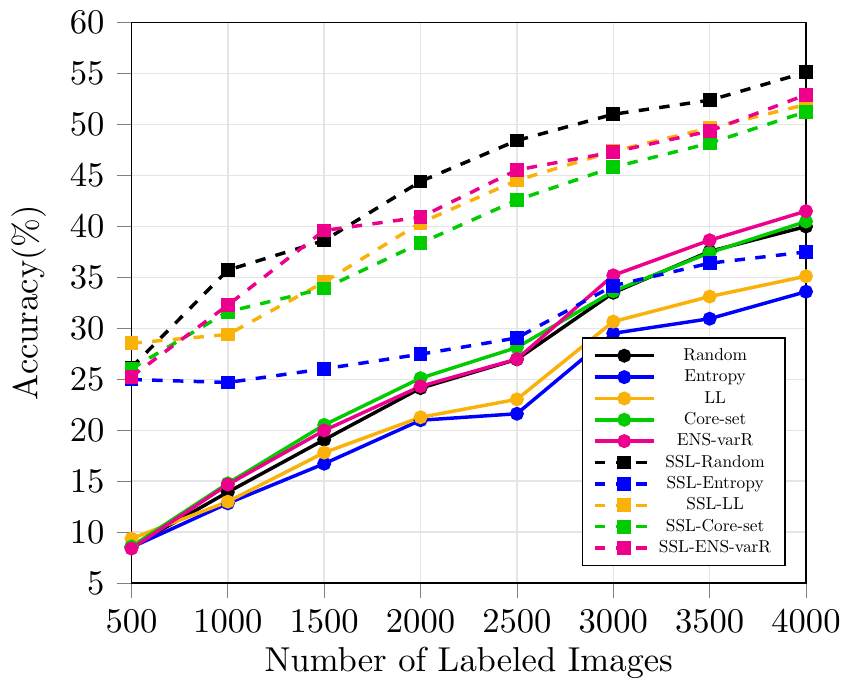}
       \caption{
       When evaluated in the low-budget regime ($\mathcal{B}_i=\mathcal{B}_s=500$) on CIFAR-100, most integrated SSL-AL methods are still better than their raw counterparts but nothing beats SSL with random sampling.}
       % Results on CIFAR-100 comparing SSL integrated active learning (SSL-X) methods with raw active learning methods. Results in low-budget setting, where $\mathcal{B}_i=\mathcal{B}_s=500$.}
    \label{fig:c100_500}
\end{figure}

The results are shown in Figures~\ref{fig:c10_250} and \ref{fig:c100_500}.
None of the active learning methods consistently outperforms the random baseline, neither on CIFAR-10 nor on CIFAR-100. This holds always for the combination of active learning and semi-supervised learning, whereas for raw active learning only ENS-varR could marginally outperform the random baseline. In fact, some techniques perform considerably worse than the random baseline, especially in conjunction with semi-supervised learning, showing that their selection strategy is counter-productive in the low-budget regime.

% ------------------ Section 3.2.5 -----------------

\subsubsection{Comparison to Transfer Learning}

Oliver \etal \cite{NIPS2018_7585} argued that transfer learning may be a preferable alternative to semi-supervised learning when a suitable labeled dataset is available for transfer learning. Following the recommendation, we compare the performance of the SSL-Random baseline with a fine-tuned ImageNet pre-trained network on CIFAR-10.

\begin{figure}[b]
\centering
    \includegraphics[width=\columnwidth]{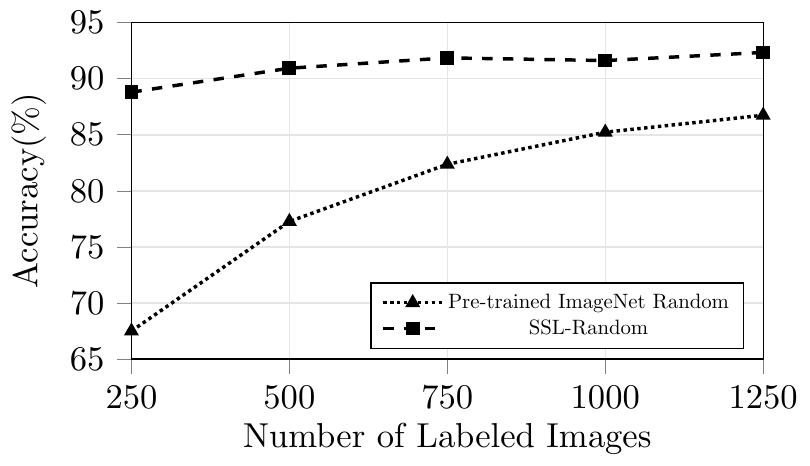}
    \caption{The SSL-Random baseline clearly outperforms a fine-tuned network pre-trained on ImageNet in the low-budget setting. Results shown on CIFAR-10.}
    % Results comparing SSL-Random baseline with the network fine-tuned over the pre-trained ImageNet weights for low-budget setting. }
    \label{fig:imagenet_250}
\end{figure}

\begin{figure}[b]
    \includegraphics[width=\columnwidth]{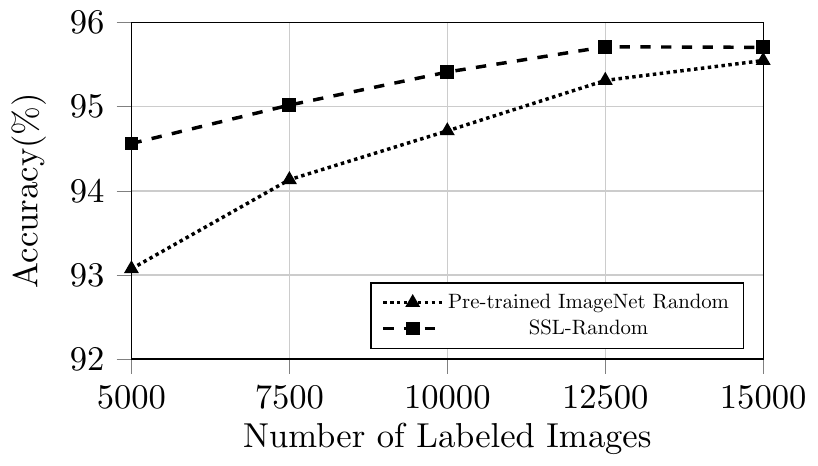}
    \caption{The SSL-Random baseline clearly outperforms a fine-tuned network pre-trained on ImageNet in the large-budget setting. Results shown on CIFAR-10.}
    % Results comparing SSL-Random baseline with the network fine-tuned over the pre-trained ImageNet weights for large-budget setting}
    \label{fig:imagenet_5k}
\end{figure}

The ImageNet pre-trained network is fine-tuned only on the labeled samples. The experiment was conducted with Resnet-18 due to the availability of pre-trained ImageNet weights. We observe that the SSL-AL method clearly outperforms fine-tuning of a pre-trained ImageNet network in both high- and low-budget settings.
We tested both budget setting for 4 sampling cycles, the corresponding results are shown in Figure \ref{fig:imagenet_250} and \ref{fig:imagenet_5k} respectively.
This experiment shows that including an up-to-date semi-supervised learning algorithm into an active learning pipeline makes sense even when large pre-training data is available.

%% file: 4_methods_SS.tex
\section{Active Learning for Semantic Segmentation} \label{Sec:SS}

Image classification is a standard active learning task.
However, with its relatively low annotation cost (1 click per image) it is not necessarily the most important one.
In this section we aim to evaluate the applicability of active learning methods to a task with a significantly higher labeling cost - semantic segmentation.
We adapt existing AL methods for classification to support semantic segmentation.

% While labeling each image for classification requires one click per image, labeling an image for dense prediction tasks like semantic segmentation costs manifold. To further validate our observations from the image classification experiments, we evaluate active learning methods for semantic segmentation. We create a few practical baseline methods spanning different types of AL methods for semantic segmentation.

%Due to unavailability of the implementation of previous works for semantic segmentation, we create a few related practical baseline methods spanning different types of AL methods.

%--------------------- Annotation Model ----------------------------

\subsection{Annotation model}

\begin{figure}
    \includegraphics[width=\columnwidth]{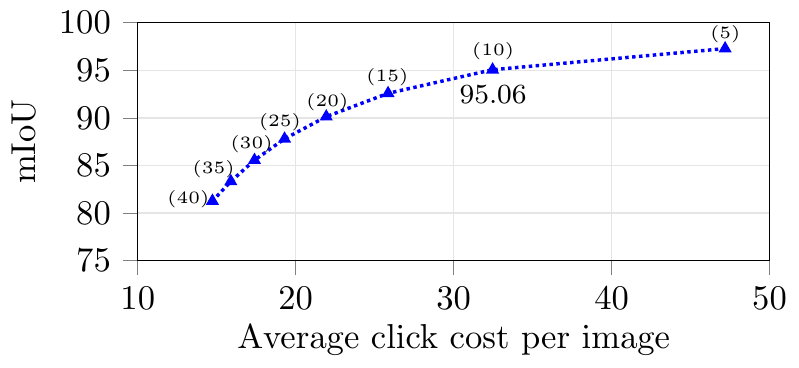}
       \caption{Trade-off between the polygon approximation quality and the annotation cost in clicks. Tolerance values used for each measurement are mentioned above the $\blacktriangle$ markers.}
    \label{fig:cost_vs_miou}
\end{figure}

\begin{figure}[b]
\begin{tabular}{c@{\hspace{1mm}}c@{\hspace{1mm}}c@{\hspace{1mm}}c@{\hspace{1mm}}c@{\hspace{1mm}}c}
Original & 5 & 10 & 15  & 20  \\[6pt]

 \includegraphics[trim={1cm 0 2cm 0}, clip, width=15mm]{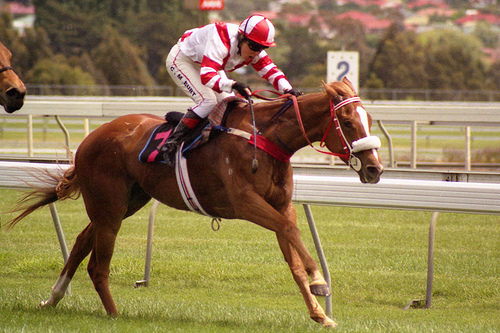} &
 \includegraphics[trim={1cm 0 3cm 0}, clip, width=15mm]{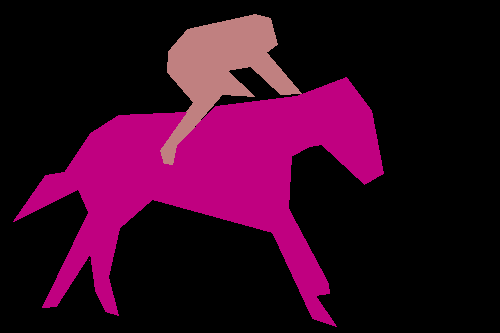} &
 \includegraphics[trim={1cm 0 3cm 0}, clip, width=15mm]{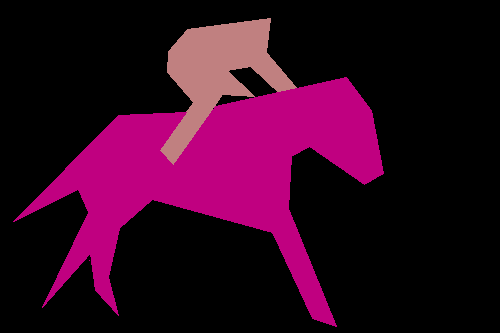} &
 \includegraphics[trim={1cm 0 3cm 0}, clip, width=15mm]{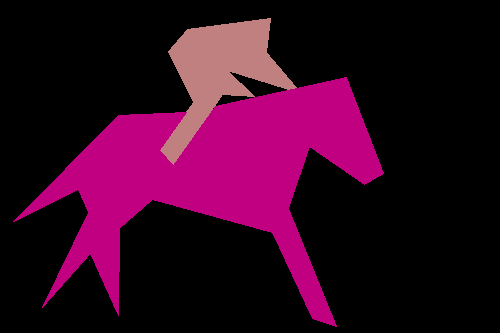} &
 \includegraphics[trim={1cm 0 3cm 0}, clip, width=15mm]{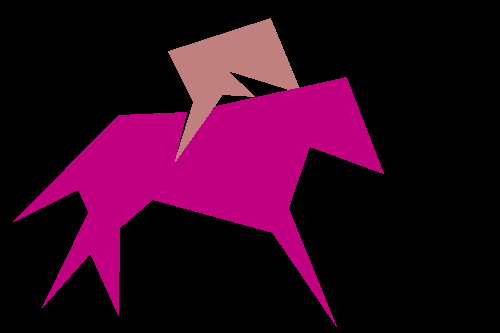} \\

 \includegraphics[trim={1cm 0 0 0}, clip, width=15mm]{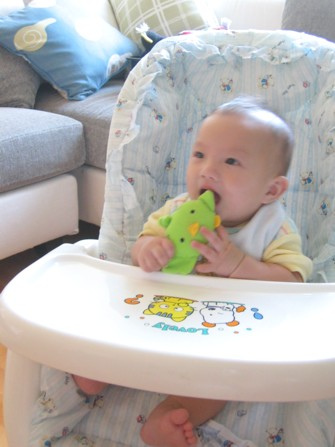} &
 \includegraphics[trim={1cm 0 0 0}, clip, width=15mm]{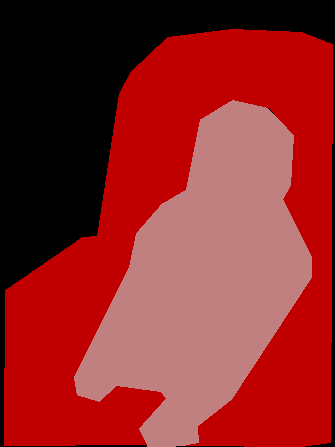} &
 \includegraphics[trim={1cm 0 0 0}, clip, width=15mm]{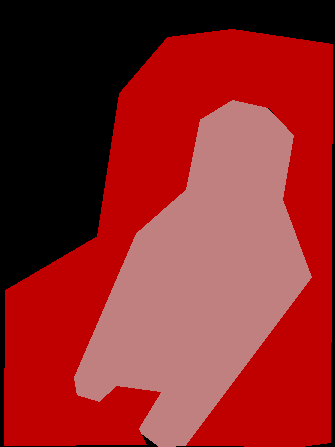} &
 \includegraphics[trim={1cm 0 0 0}, clip, width=15mm]{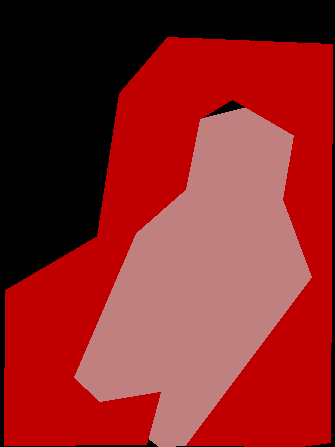} &
 \includegraphics[trim={1cm 0 0 0}, clip, width=15mm]{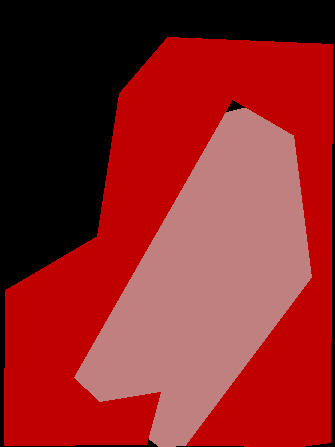} \\

\end{tabular}
\caption{Labeling quality when using polygon approximation with different tolerance values (in pixels). We picked the tolerance value of 10 pixels for our experiments.}
% Figure shows different labeling quality with polygon approximations. Columns 2-5 show labels with pixel-level labeling tolerance of 5,10,15,20 pixels respectively. We select tolerance value of 10 pixels for our experiments.}
\label{fig:tol_vs_qual}
\end{figure}

A conventional active learning setup includes a human in the loop who annotates the samples picked by the query function.
Since training with an actual human annotator is prohibitively expensive, we simulated its actions during training.
We used the number of clicks required to annotate the entire image as a proxy for the annotation cost.
To do so, we approximate each connected component in the ground truth image with a polygon using the Ramer-Douglas-Peucker algorithm \cite{Douglas2011Algorithms}.
The approximation quality is controlled by a pre-defined pixel-level tolerance parameter.
The total number of clicks per image is then calculated by adding up the number of vertices for all polygons in this image.
%Using grid search over cost vs performance, (more information in the supplementary), we select the pixel-level labeling tolerance of 10 pixels. 
 We perform a grid search over different tolerance values ranging from 5 to 40 pixels to find a suitable value. 
Figure \ref{fig:cost_vs_miou} shows the trade-off between the average click cost per image and the polygon approximation quality of annotations for different tolerance values. The trade-off between different tolerance values and labeling quality is shown in Figure \ref{fig:tol_vs_qual}.
We select the pixel-level labeling tolerance of 10 pixels. The approximated labels retain 95.06\% mIoU as compared to the original ground-truth labels. According to this approximation, an average image costs around 33 clicks to label.

%The amount of permissible noise for the annotation simulator is controlled by tolerance parameter of the Douglas-Peucker algorithm. The tolerance parameter defines the allowed pixel displacement error when a region is approximated using a polygon. We perform grid search over different tolerance values ranging from 5 to 40 pixels to find a suitable value. Figure \ref{fig:cost_vs_miou} shows the trade-off between the average click cost per image and the polygon approximation quality of annotations for different tolerance values. Finally, we select a pixel tolerance value of 10 pixels, which retains 95.06\% of the original ground-truth label performance. According to this approximation, an average image costs around 33 clicks to label. 

\subsection{Baseline Methods}
We evaluated two kinds of AL methods: uncertainty-based methods and learning-based methods. 

\mypara{Random}. We consider the random sampling method as our first baseline.

\mypara{Entropy}. This uncertainty method is based on the softmax-entropy of the segmentation prediction.
Different to the classification case, we need some integral uncertainty measure that aggregates per-pixel uncertainty values.
There are a few heuristics for this \cite{Aghdam_2019_ICCV, SA}, but none of them is considered standard.
We evaluated several simple heuristics including averaging and taking the maximum value over the image, and concluded that the best results are achieved by counting the number of pixels per-image with an uncertainty value higher than a certain threshold.
We use 0.6 as a threshold value, which was determined via grid search.
% Finally, to find a proxy uncertainty measure for the whole image, we convert entropy image into a binary mask using a threshold, such that it contains only high entropy pixels. Further, we use the area covered by these high entropy pixels as a uncertainty measure for the image. Using coarse grid search, we find the best entropy threshold value of 0.6 where the range of values lies between 0 and 1.

\mypara{Ensemble with Average Entropy (ENS-ent)}. This second uncertainty-based method ENS-ent is based on the average entropy over the predictions from all members of the model ensemble. We used the same information accumulation heuristic as used for the Entropy method.

\mypara{Learn Loss (LL)}. We adapted the LL \cite{Yoo_2019_CVPR} method from image classification to semantic segmentation. Since the original module is proposed for a resnet architecture and the segmentation network used in this work is also based on a resnet architecture, the exact method is directly adapted by reusing the original loss prediction module. 
% It was originally proposed for image classification.
% The implementation setup of LL method for segmentation is similar to image classification setup.

\begin{figure*}[h!]
\begin{tabular}{c@{\hspace{1mm}}c@{\hspace{1mm}}c@{\hspace{1mm}}c@{\hspace{1mm}}c@{\hspace{1mm}}c@{\hspace{1mm}}c@{\hspace{1mm}}c}
 Original & 5k & 10k & 15k  & 20k & 25k & 30k & GT \\[6pt]
 
 \begin{overpic}[trim={1cm 0 0 0}, clip, width=20mm]{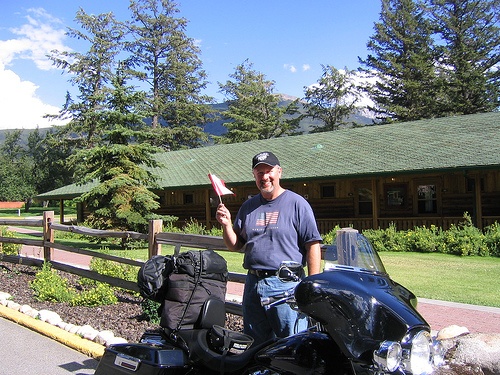}
\put (-10,8.0) {\color{black}\footnotesize \rotatebox{90}{Entropy}}
\end{overpic} &

 \includegraphics[trim={1cm 0 0 0}, clip, width=20mm]{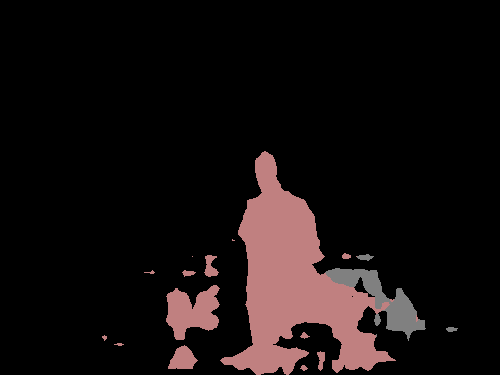} &
 \includegraphics[trim={1cm 0 0 0}, clip, width=20mm]{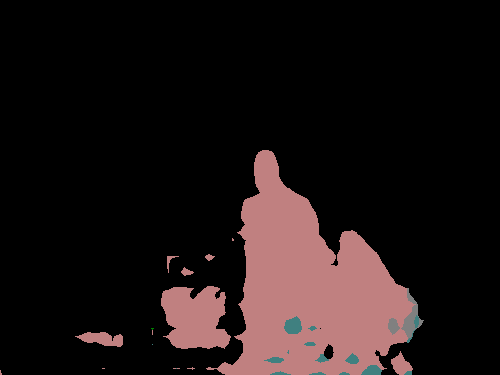} &
 \includegraphics[trim={1cm 0 0 0}, clip, width=20mm]{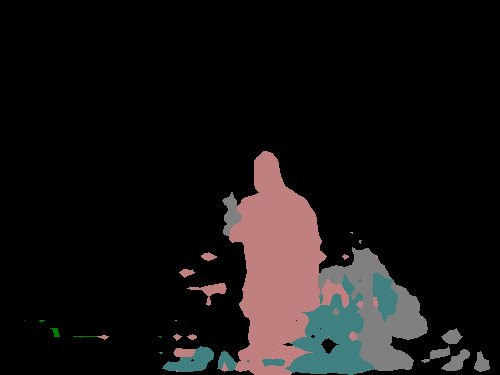} &
 \includegraphics[trim={1cm 0 0 0}, clip, width=20mm]{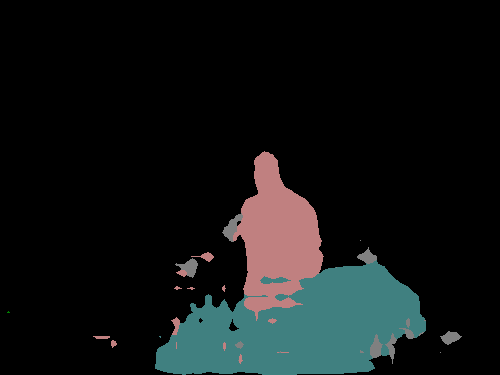} &
 \includegraphics[trim={1cm 0 0 0}, clip, width=20mm]{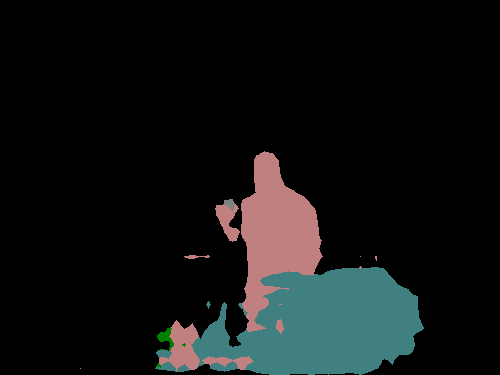} &
 \includegraphics[trim={1cm 0 0 0}, clip, width=20mm]{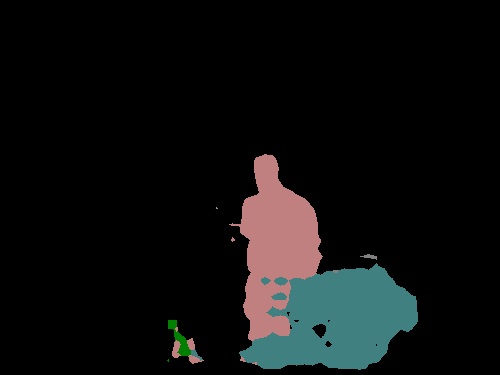} & 
 \includegraphics[trim={1cm 0 0 0}, clip, width=20mm]{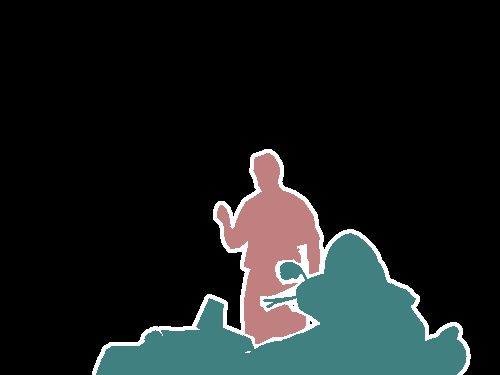} \\
 
  \begin{overpic}[trim={1cm 0 0 0}, clip, width=20mm]{./results/polygon_results/2007_002823.jpg}
\put (-10,1.0) {\color{black}\footnotesize \rotatebox{90}{SSL-Random}}
\end{overpic} &

 \includegraphics[trim={1cm 0 0 0}, clip, width=20mm]{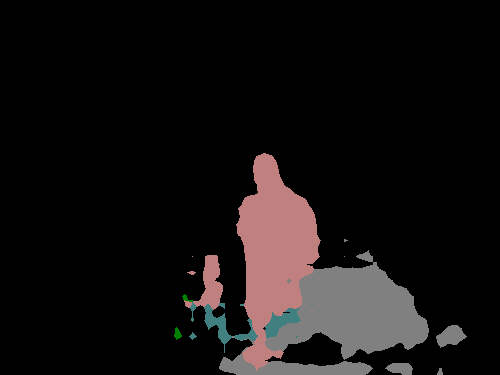} &
 \includegraphics[trim={1cm 0 0 0}, clip, width=20mm]{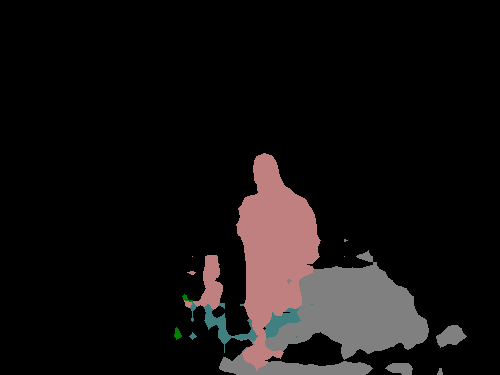} &
 \includegraphics[trim={1cm 0 0 0}, clip, width=20mm]{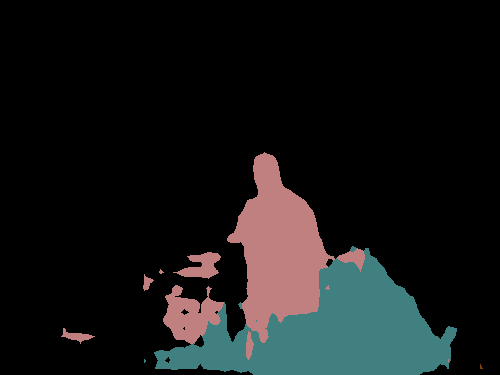} &
 \includegraphics[trim={1cm 0 0 0}, clip, width=20mm]{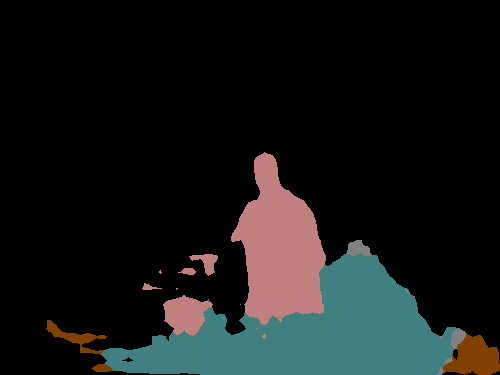} &
 \includegraphics[trim={1cm 0 0 0}, clip, width=20mm]{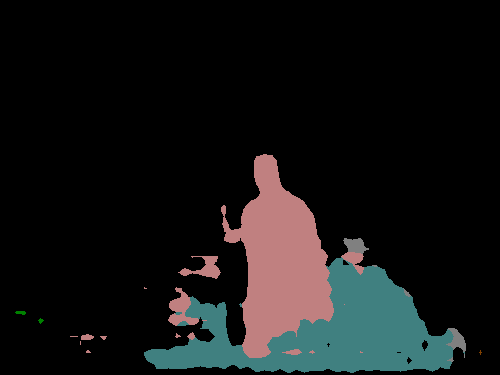} &
 \includegraphics[trim={1cm 0 0 0}, clip, width=20mm]{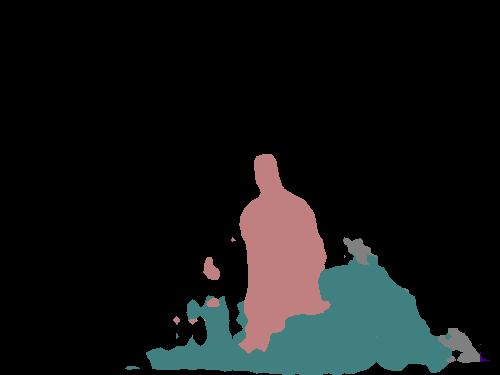} &
  \includegraphics[trim={1cm 0 0 0}, clip, width=20mm]{./results/polygon_results/2007_002823_gt.png} \\
  \begin{overpic}[trim={1cm 0 0 0}, clip, width=20mm]{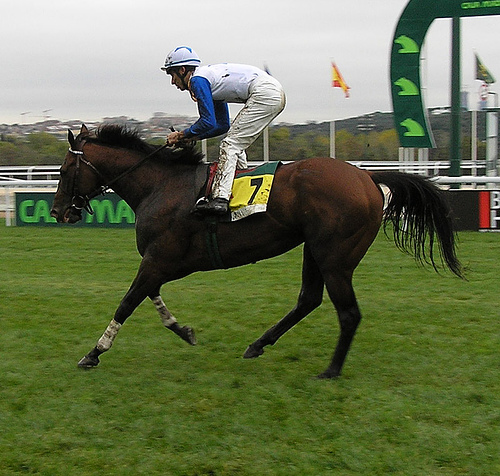}
\put (-10,16.0) {\color{black}\footnotesize \rotatebox{90}{Entropy}}
\end{overpic} &
 \includegraphics[trim={1cm 0 0 0}, clip, width=20mm]{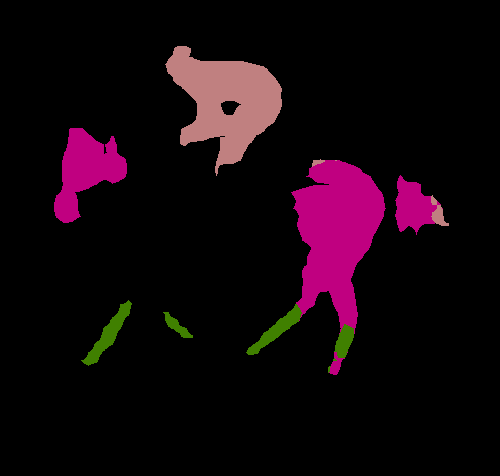} &
 \includegraphics[trim={1cm 0 0 0}, clip, width=20mm]{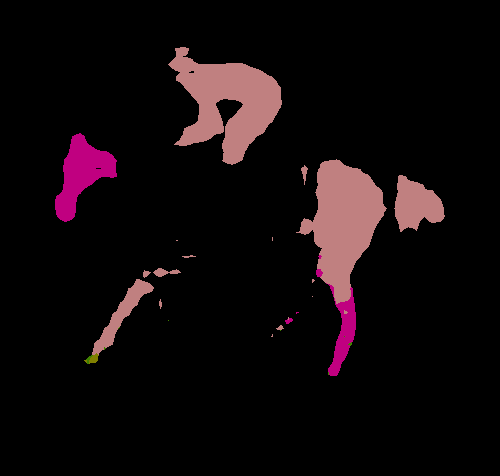} &
 \includegraphics[trim={1cm 0 0 0}, clip, width=20mm]{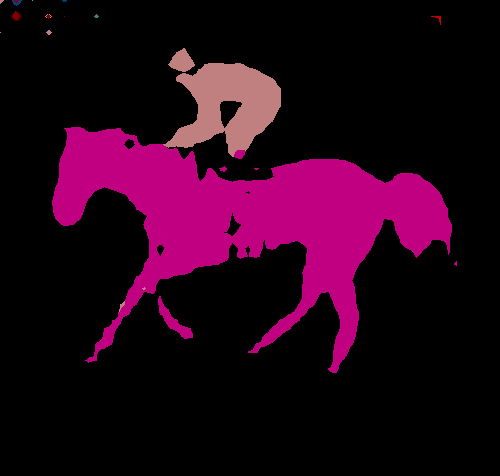} &
 \includegraphics[trim={1cm 0 0 0}, clip, width=20mm]{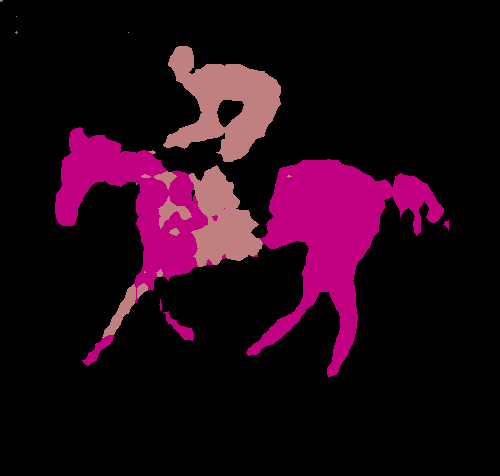} &
 \includegraphics[trim={1cm 0 0 0}, clip, width=20mm]{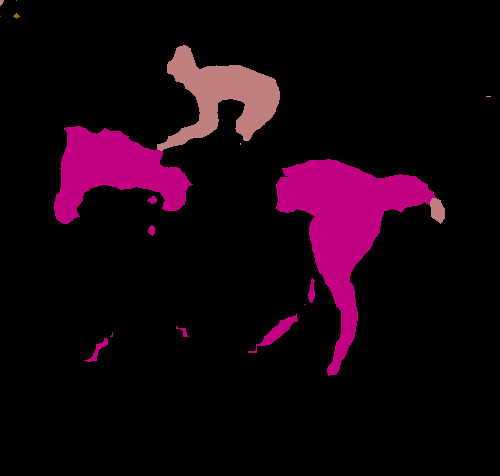} &
 \includegraphics[trim={1cm 0 0 0}, clip, width=20mm]{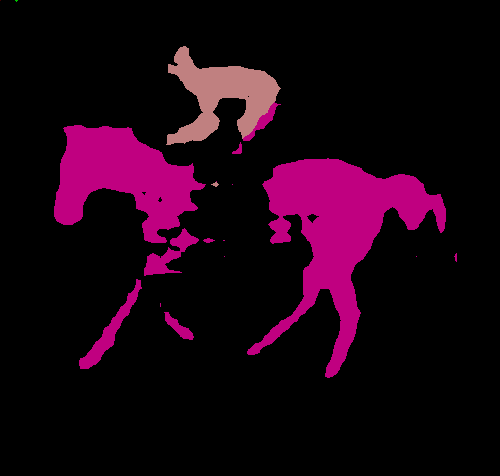} & 
 \includegraphics[trim={1cm 0 0 0}, clip, width=20mm]{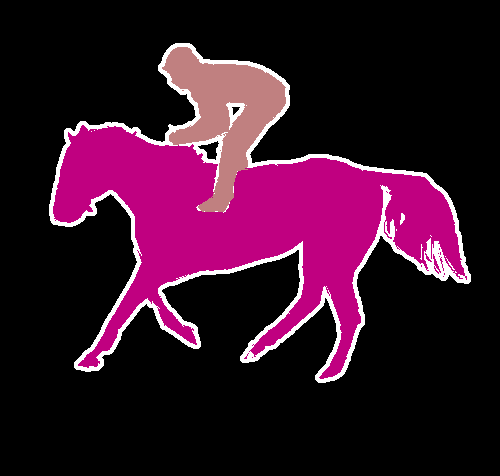} \\
 
 \begin{overpic}[trim={1cm 0 0 0}, clip, width=20mm]{./results/polygon_results/2010_000038.jpg}
\put (-10,7.0) {\color{black}\footnotesize \rotatebox{90}{SSL-Random}}
\end{overpic} &
 \includegraphics[trim={1cm 0 0 0}, clip, width=20mm]{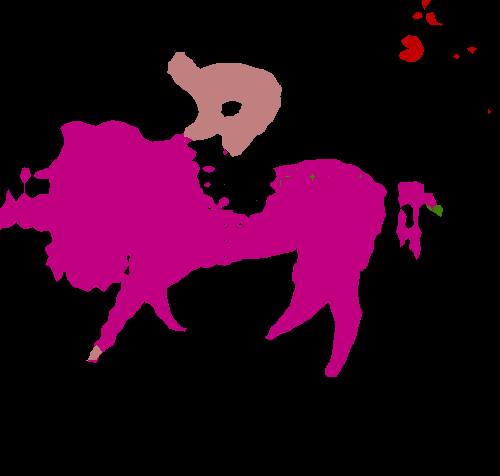} &
 \includegraphics[trim={1cm 0 0 0}, clip, width=20mm]{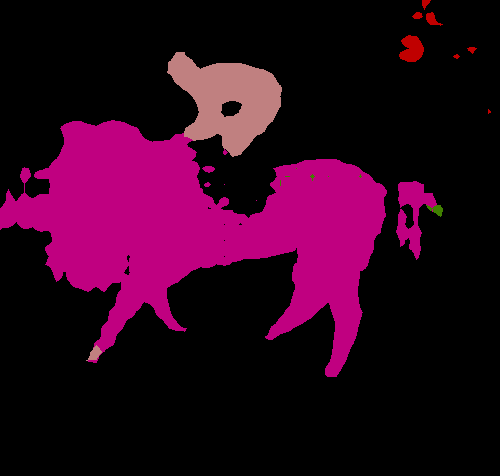} &
 \includegraphics[trim={1cm 0 0 0}, clip, width=20mm]{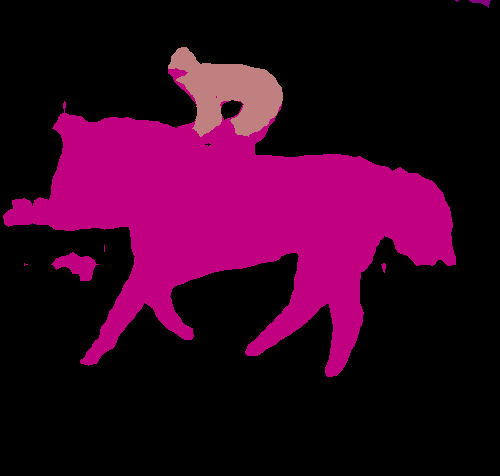} &
 \includegraphics[trim={1cm 0 0 0}, clip, width=20mm]{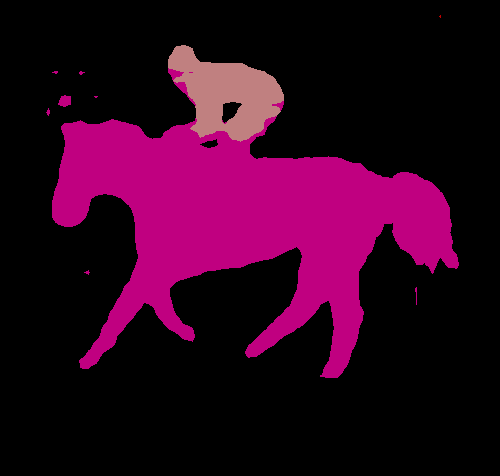} &
 \includegraphics[trim={1cm 0 0 0}, clip, width=20mm]{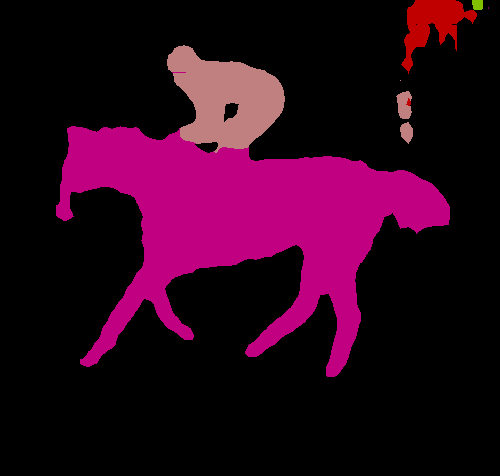} &
 \includegraphics[trim={1cm 0 0 0}, clip, width=20mm]{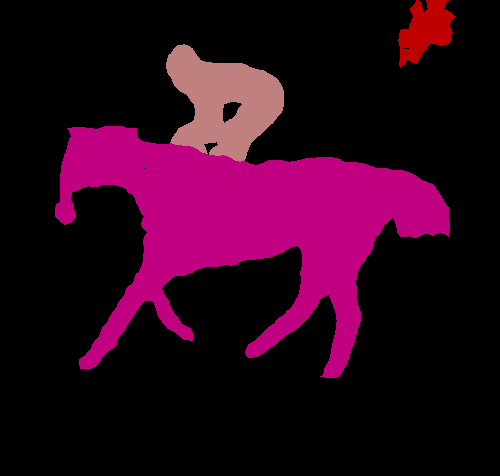} &
  \includegraphics[trim={1cm 0 0 0}, clip, width=20mm]{./results/polygon_results/2010_000038_gt.png} \\

\end{tabular}
\caption{Qualitative semantic segmentation results on PASCAL-VOC at each cycle comparing the Entropy-Image method and the SSL-Random-Image baseline. The column headings indicate the click budget used to train the corresponding model.}
\label{fig:voc_cycles}
\end{figure*}

\mypara{Semi-supervised Learning (s4GAN).}
To leverage unlabeled samples, we used the semi-supervised semantic segmentation method by Mittal \etal \cite{1908.05724}. It has been shown to produce large performance gain with as few as 2\% labeled samples on the PASCAL VOC dataset.
We only used the s4GAN branch of the proposed SSL method, which can be trained in an end-to-end manner, and dropped the classification branch for simplicity. 
%We use s4GAN branch because majority of the improvement comes from this branch and it is simple to implement. 
The s4GAN method is based on a conditional generative adversarial network, which uses the segmentation network as a generator. The discriminator of the s4GAN discriminates between the predicted and ground-truth segmentation masks. 
We used the same hyper-parameters as provided by Mittal \etal~\cite{1908.05724} for our experiments. We also combine all the above mentioned AL methods with the s4GAN method and evaluate them in the active learning setting. 
 
\mypara{SSL-D-score.}
Inspired by Ravanbakhsh~\etal~\cite{ravanbakhsh:MIDLAbstract2019a}, we propose to use the discriminator of the s4GAN network as a query function for sampling. 
The output of the discriminator varies between 0 and 1, where higher score is assigned to a higher quality of segmentation prediction. 
%0 is assigned to the predicted segmentation maps and 1 to the ground-truth segmentation maps. 
In other words, the discriminator of the s4GAN network acts as a critic which provides a higher rating for better segmentation quality. This heuristic selects the samples which are not well represented by the current learned model, which is indicated with lower rating. We refer to this semi-supervised approach for active learning as the SSL-D-score method. %\thomas{Paragraph ok, but could read better. Too complicated phrases.} %\sudhanshu{now?}

% ------------- Experiments and Results ------------

\subsection{Experimental design}

We show the performance of the AL methods for semantic segmentation on PASCAL-VOC 2012 \cite{pascal_voc}. The dataset consists of 20 foreground classes and one background class. We use the augmented annotated dataset which contains 10582 training images and 1449 validation images.

\mypara{Data Setting}. In AL experiments for segmentation, we define the labeling cost in clicks. We use the initial labeling budget $\mathcal{B}_i$ and subsequent sampling budget $\mathcal{B}_s$ of 5000 clicks, which is approximately 1.5\% of total labeling cost of the dataset. In the first cycle, randomly sampled images are completely labeled until $\mathcal{B}_i$ is exhausted. In the subsequent cycles, an AL query method selects images based on a certain criterion and labels the picked image until $\mathcal{B}_s$ is exhausted. We test all the segmentation AL methods for 5 sampling cycles. All the results are shown on the validation set.

\mypara{Training Details}. We used the DeepLabv2 \cite{DBLP:journals/pami/ChenPKMY18} architecture for all the experiments. The DeepLabv2 model is pre-trained using Microsoft COCO \cite{mscoco} dataset. 
The network is optimized using a SGD optimizer with a base-learning rate of 2.5e-4, momentum of 0.9 and a weight decay of 5e-4. We use poly-learning policy similar to the original segmentation work \cite{DBLP:journals/pami/ChenPKMY18}. We use a batch size of 10 and train each cycle for 150 epochs. The network gets an input image of size $321 \times 321$.
The used DeepLabv2 version achieves a performance of 73.6 and 71.6 mIoU on original and approximated ground-truth labels respectively on the validation set. 
The combined semi-supervised active learning (SSL-AL) methods are trained for 10k iterations for each cycle with a batch size of 8. The training procedure and hyperparameters for semi-supervised learning are the same as in \cite{1908.05724}.

We evaluate AL methods for semantic segmentation in two different settings: (1) using standard augmentations and (2) using semi-supervised learning for training the target model. The mean performance is reported over 3 trials for all single model-based methods and over 2 trials for ensemble-based methods due to higher computation cost.

\subsection{Results} \label{Sec:SS_SSL}

\subsubsection{Is active learning beneficial in more labor-intense labeling tasks than image classification, e.g. semantic segmentation?}

%\thomas{Combine the two settings into one subsection? } \sudhanshu{Yeah, makes sense. Done}

\begin{figure}[t]
    \includegraphics[width=\columnwidth]{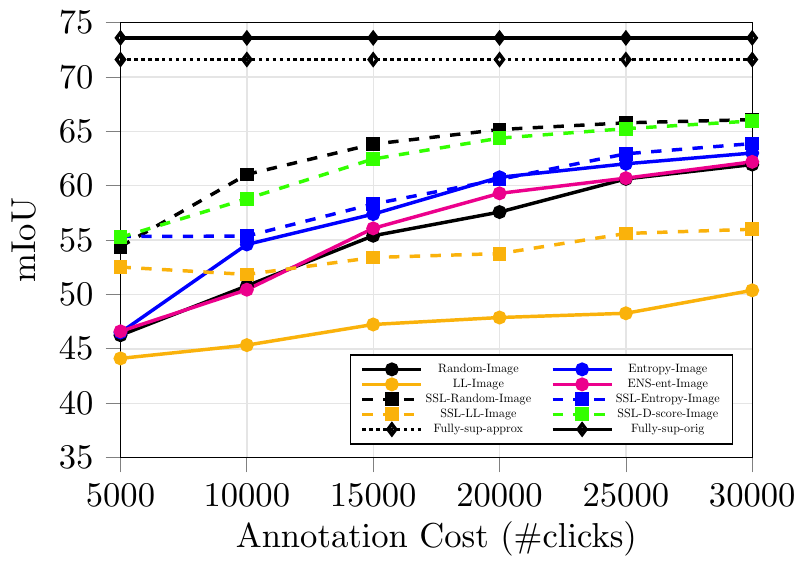}
       \caption{Integrated SSL-AL methods for semantic segmentation mostly perform better than their raw counterparts on PASCAL-VOC with $\mathcal{B}_i=\mathcal{B}_s=5000$ clicks ($\approx1.5\%$ of the dataset). None of the methods outperforms SSL with random sampling.}
       % Results for semantic segmentation comparing SSL integration with active learning (SSL-X) over the standard setting. Results shown on PASCAL-VOC dataset, where $\mathcal{B}_i=\mathcal{B}_s=5000$ clicks ($\approx1.5\%$ of dataset). The suffix `Image' refers to the image-level labeling regime. }
    \label{fig:voc_image}
\end{figure}

%\subsubsection{Standard Setting} \label{asl_aug}
To find the relevance of active learning for semantic segmentation, we first train the model only using augmentations including random horizontal flipping and random resized cropping. Other geometry preserving augmentation like brightness, contrast, rotation, scaling do not show any measurable improvement for semantic segmentation \cite{anonymous2020semisupervised}. In the results, the uncertainty method based on entropy performs best and shows an improvement of around 1.1\% mIoU over the random sampling baseline after 5 AL sampling cycles. The LL method fails to outperform the random baseline approach. Corresponding performance curves are shown with solid lines in Figure \ref{fig:voc_image}. 
%\ozgun{is pink dashed missing?}

%\subsubsection{Using Unlabeled Pool for Training} \label{Sec:SS_SSL}

\begin{figure}[b]
\begin{tabular}{c@{\hspace{1mm}}c@{\hspace{1mm}}c@{\hspace{1mm}}c@{\hspace{1mm}}c@{\hspace{1mm}}c}

 \includegraphics[trim={1cm 0 0 0}, clip, width=20mm]{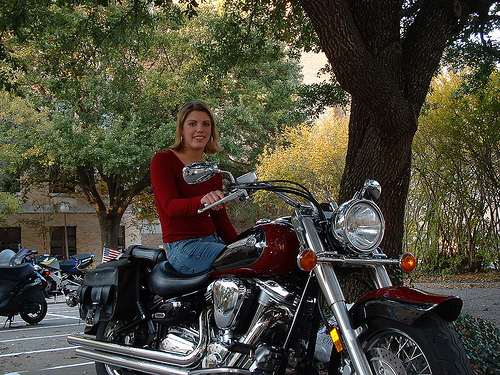} &
 \includegraphics[trim={1cm 0 0 0}, clip, width=20mm]{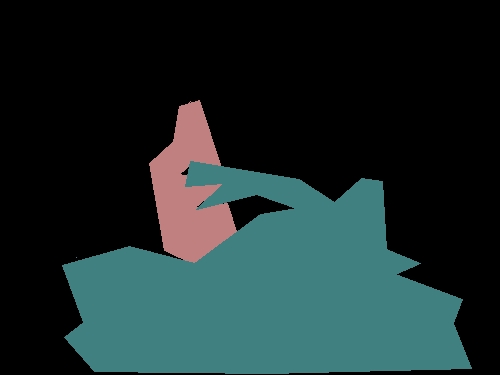} &
 \includegraphics[trim={1cm 0 0 0}, clip, width=20mm]{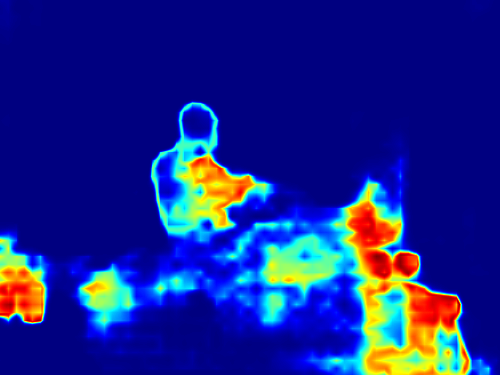} &
 \frame{\includegraphics[trim={1cm 0 0 0}, clip, width=20mm]{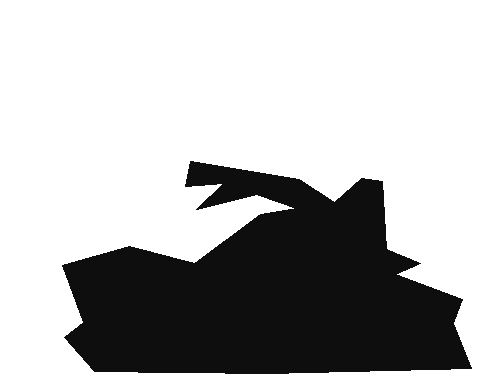}} \\
 
  \includegraphics[trim={1cm 0 0 0}, clip, width=20mm]{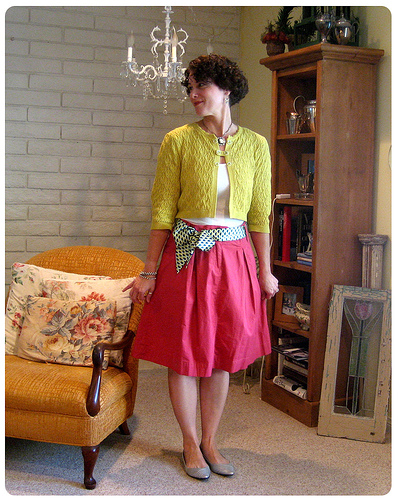} &
 \includegraphics[trim={1cm 0 0 0}, clip, width=20mm]{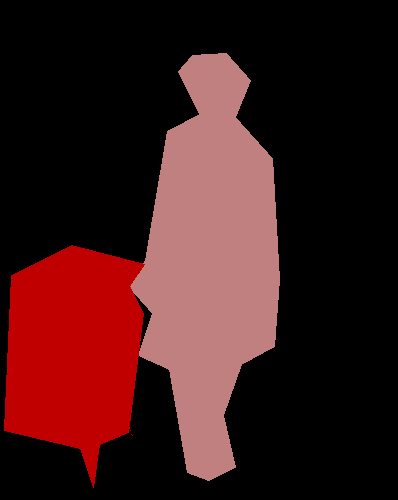} &
 \includegraphics[trim={1cm 0 0 0}, clip, width=20mm]{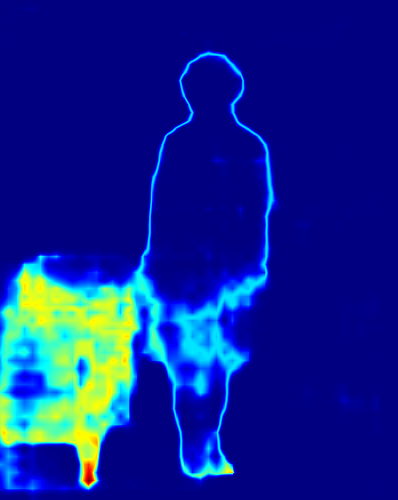} &
 \frame{\includegraphics[trim={1cm 0 0 0}, clip, width=20mm]{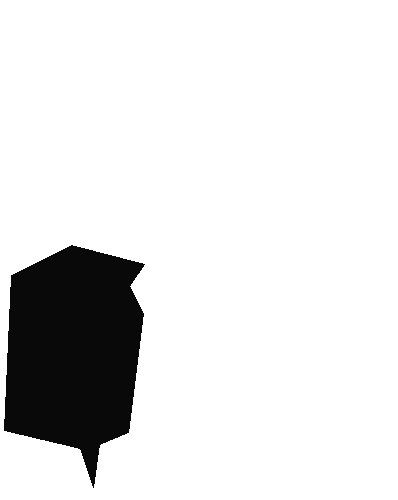}}\\
\end{tabular}
\caption{Image labeling in a polygon-level labeling regime. From left to right: Original image, approximated ground-truth, pixel-wise entropy and selected polygon for labeling based on the entropy heuristic.}
\label{fig:entropy_polygon}
\end{figure}

Further, we use the unlabeled pool of images to train the target model for semantic segmentation and analyze the performance. We utilize the s4GAN model \cite{1908.05724} to leverage the information from unlabeled samples. When used on top of the SSL method, all AL methods show a clear gain in performance. The performance of the random baseline (Random) when combined with s4GAN increases by the largest margin of 4.1\% mIoU and reaches the overall best value. In addition, SSL-D-score heuristic also shows comparable performance to the random baseline after 5 sampling cycles, but does not bring any improvement over the SSL-Random baseline. The performance curves for all integrated methods are shown with dashed lines in Figure \ref{fig:voc_image}. 
Figure \ref{fig:voc_cycles} shows the qualitative results at each sampling cycle, comparing the Entropy-Image method and the SSL-Random-Image baseline. 

\begin{figure}[b]
    \includegraphics[width=\columnwidth]{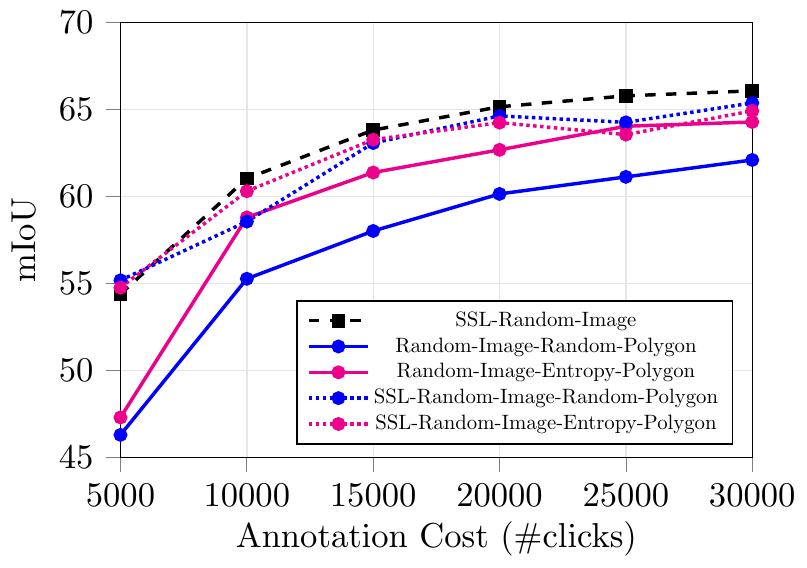}
    \caption{Active learning for semantic segmentation: comparison between SSL integrated with active learning (SSL-X) against the standard setting. Results shown on the PASCAL-VOC dataset with $\mathcal{B}_i=\mathcal{B}_s=5000$ clicks. The suffixes `Image' and `Polygon' refer to the image-level and polygon-level labeling regimes respectively.}
    \label{fig:voc_polygon}
\end{figure}

\subsubsection{Polygon-level Labeling}

Here, we explore whether labeling only a part of an image is more effective than labeling the complete image.
We evaluate active learning methods for semantic segmentation, where only a region of an image is selected by the query function. This region is approximately labeled using a polygon by the annotation simulator. We evaluate methods where an image is selected randomly, but the polygon in the image is selected based on the active learning heuristic. We compare entropy-based and random polygon selection methods in both raw and SSL-integrated active learning settings.

\mypara{Experiment Details.} Entropy of a polygon is measured in a similar way as in the image-level labeling regime. We create a binary mask for the pixel-wise entropy based on a threshold and use the area of the high-entropy pixels as our selection heuristic. Only in the first cycle, images are completely labeled until the $\mathcal{B}_i$ is covered. In the subsequent cycles, images are labeled polygon-wise until the sampling budget $\mathcal{B}_s$ is exhausted. Figure \ref{fig:entropy_polygon} shows two examples of how polygon-level labeling regime works based on the entropy heuristic. The budget settings and the hyperparameters exactly match those from the image-level labeling regime. %This method is able to label more images under the same budget but only partially. We found that the best image-level labeling method is by 0.7\% mIoU points better than the best region-level labeling method.

\mypara{Results.} Entropy-based polygon selection approach is more effective than random polygon selection for the raw active learning (without SSL) setting. However, when combined with semi-supervised learning, both entropy (Random-Image-Entropy-Polygon) and random (Random-Image-Random-Polygon) polygon selection strategies perform very similarly. Results are shown in Figure \ref{fig:voc_polygon}. Moreover when all polygon-level labeling approaches are compared with the image-level labeling approaches, we find SSL-Random-Image baseline even outperforms all the polygon-level active learning methods.

In this experiment, we also observed that the SSL-Random-Image baseline outperformed the SSL-Random-Image-Random-Polygon baseline, showing that image-level labeling is a more effective way of labeling an image.

%These observations also align with our previous conclusions: Making active learning methods work for semantic segmentation is important but not trivial.

%We also tested region-level labeling where only a polygon of the image is labeled based on a heuristic, instead of labeling the entire image. This method is able to label more images under the same budget but only partially. We found that the best image-level labeling method is by 0.7\% mIoU points better than the best region-level labeling method. Detailed results for the region-level labeling experiment can be found in the supplementary.

%% file: 5_discussion.tex
\section{Discussion}

Our experiments provide strong evidence that the current evaluation protocol used in active learning is sub-optimal which in turn leads to wrong conclusions about the methods' performance and the state of the field in general.

%Evaluating on a dataset marginally different from the one conventionally used -- CIFAR-100 instead of CIFAR-10 -- dramatically changes the ranking of the methods. 
Evaluating on CIFAR-100 which is marginally different from CIFAR-10, dramatically changes the ranking of the methods. 
Applying state-of-the-art data augmentation significantly increases the scores of all methods making them virtually indistinguishable in terms of final performance.

Modern semi-supervised learning algorithms applied in the conventional active learning setting show a higher relative performance increase than any of the active learning methods proposed in the recent years.

State-of-the-art active learning approaches often fail to outperform simple random sampling, especially when the labeling budget is small - a setting critically important for many real-world applications.

Based on these observations, we formulate a more appropriate evaluation protocol and recommend using it for benchmarking future active learning methods.

\begin{enumerate}
\item AL methods should be evaluated on a wider range of datasets to assess their general robustness.

\item It is important to evaluate AL methods with up-to-date network architectures and up-to-date augmentation techniques.

\item There should always be a direct comparison between AL methods and SSL methods.

\item Together with the existing large-budget regime, AL methods should be evaluated in the low-budget regime.
\end{enumerate}

%In addition, our results indicate that a combination of SSL and AL, while beneficial in some cases, is not robust. We believe that reliably combining the two objectives is an important research direction. Also, we find that the classification task with its low annotation cost should not be the only testbed for developing active learning methods. Tasks with higher labeling cost can potentially benefit much more from employing informed sample selection strategies.

%\sudhanshu{Should we discuss possible reasons for underperformance of AL methods (raw and integrated), especially in low-budget setting?}

It would be interesting to know why AL often performs worse than random sampling and consistently does so in the low-budget regime. For now, we can only speculate. We believe that AL sampling introduces a bias into the distribution of annotated samples, i.e., the sampled distribution does not sufficiently match the true distribution anymore. The damage by this bias is larger than the positive effect of learning from ``more interesting'' samples. If this hypothesis is true, research in active learning should focus on ways that avoid any bias by the selection strategy. 

Our results also indicate that adapting AL methods to tasks with higher labeling cost, e.g. semantic segmentation, is a non-trivial problem.
Although there is not enough empirical evidence for this, we speculate that such tasks can potentially benefit much more from employing informed sample selection strategies and thus define a promising direction for future research.

%In fact, though being even more speculative, the relatively good performance on CIFAR-10, which breaks down on other datasets, could be explained by researchers tweaking their methods to manually remove some of the bias specifically for this dataset. If this tweaking could be turned into a concept that generalizes across datasets, there is again hope for a consistent added value by active learning. 